\documentclass[a4paper,11pt]{article}
\usepackage[margin=1in]{geometry}

\usepackage{amsmath,amsfonts,bm}

\def\eqref#1{equation~\ref{#1}}

\def\1{\bm{1}}

\DeclareMathAlphabet{\mathsfit}{\encodingdefault}{\sfdefault}{m}{sl}
\SetMathAlphabet{\mathsfit}{bold}{\encodingdefault}{\sfdefault}{bx}{n}

\def\C{{\mathbb{C}}}

\usepackage{graphicx}
\usepackage{natbib}
\usepackage{booktabs}
\usepackage{xcolor}
\usepackage{url}
\usepackage{parskip}
\usepackage[colorlinks = true,
            linkcolor = blue,
            urlcolor  = blue,
            citecolor = blue,
            anchorcolor = blue]{hyperref}

\usepackage{cleveref}
\usepackage{multirow} 
\usepackage{array}

\title{A Deep State Space Model for Rainfall-Runoff Simulations}

\author{Yihan Wang$^{1}$ \,, \,\, Lujun Zhang$^{1}$, \,\, Annan Yu$^{2}$, \,\, N. Benjamin Erichson$^{3}$, \\ and Tiantian Yang$^{1}$\thanks{Corresponding author: \url{tiantian.yang@ou.edu}} \vspace{+0.3cm} \\ 
$^1$ School of Civil Engineering and Environmental Science, University of Oklahoma \\
$^2$ Center for Applied Mathematics, Cornell University \\
$^3$ Lawrence Berkeley National Laboratory \\
}

\begin{document}

\maketitle


\begin{abstract}

The classical way of studying the rainfall-runoff processes in the water cycle relies on conceptual or physically-based hydrologic models. Deep learning (DL) has recently emerged as an alternative and blossomed in hydrology community for rainfall-runoff simulations. However, the decades-old Long Short-Term Memory (LSTM) network remains the benchmark for this task, outperforming newer architectures like Transformers. In this work, we propose a State Space Model (SSM), specifically the Frequency Tuned Diagonal State Space Sequence (S4D-FT) model, for rainfall-runoff simulations. The proposed S4D-FT is benchmarked against the established LSTM and a physically-based Sacramento Soil Moisture Accounting model across 531 watersheds in the contiguous United States (CONUS). Results show that S4D-FT is able to outperform the LSTM model across diverse regions. Our pioneering introduction of the S4D-FT for rainfall-runoff simulations challenges the dominance of LSTM in the hydrology community and expands the arsenal of DL tools available for hydrological modeling.

\end{abstract}


\section{Introduction}
\label{sec1}
The rainfall-runoff relationship is a fundamental concept in hydrology. It describes how rainfall is transformed into surface runoff through interconnected hydrologic processes, such as infiltration, evapotranspiration, and the exchange of water between surface and subsurface flows 
 \citep{beven1979physically}. Thoroughly understanding these hydrologic processes and subsequently achieving accurate simulations of the rainfall-runoff relationship are critical for proactive flood forecasting and mitigation, efficient agricultural planning, and strategic urban development  \citep{beven2012rainfall,knapp1991review,moradkhani2008general}.

Physically-based hydrologic models (PBMs), grounded in physical laws that govern hydrologic dynamics, are the standard tools for simulating rainfall-runoff relationships \citep{beven1996discussion}. However, the highly nonlinear nature of various hydrologic processes often challenges PBMs, limiting their accuracy in diverse conditions  \citep{beven1989changing,clark2017evolution}. Consequently, there is a growing need for innovative approaches to address the limitations of PBMs. 

Deep learning (DL) has emerged as an alternative to PBMs, showing success in capturing the complex, nonlinear patterns in rainfall-runoff simulations. The hydrology community also explores the applicability of DL models in rainfall-runoff simulations across diverse temporal scales and geospatial locations. For the large-scale studies that focus the model evaluation in the contiguous United States (CONUS), it is recognized that the decade-old Long Short-Term Memory (LSTM) networks \citep{hochreiter1997long} continue to be the best-performing architecture for rainfall-runoff simulations, with even Transformers \citep{vaswani2017attention} unable to outperform LSTMs \citep{frame2022deep,kratzert2019toward,kratzert2019towards,liu2024probing}.

In this work, we pioneer the uses of a new set of State Space Models (SSMs) ~\citep{gu2021efficiently, gu2021combining} for rainfall-runoff simulations. Since the original invention, the SSMs have achieved state-of-the-art performance across diverse tasks in video, audio, and time-series processing, excelling at long-range sequence modeling while being faster and more memory-efficient than LSTMs and Transformers \citep{gu2023mamba,patro2024mamba}. However, to the best of the authors’ knowledge, there is no single study that tested out SSMs in hydrologic simulations and studied their capabilities in simulating the rainfall-runoff processes in the field of hydrology. Therefore, in this work, we raise the scientific question: Can SSMs enhance rainfall-runoff simulations and outperform the decades-old LSTM model that has been extensively used and widely accepted as the DL benchmark in hydrology community? To answer this question, we evaluate the Frequency Tuned Diagonal State Space Sequence (S4D-FT) model ~\citep{yu2024tuning} for rainfall-runoff simulations across 531 watersheds in CONUS and carry out a comprehensive evaluation of the model performance. Our evaluation follows a three-step approach. Firstly, the overall statistical performance of S4D-FT is compared with the basic S4D (without frequency tuning)~\citep{gu2022parameterization}, various existing DL benchmarks, and a traditional PBM (Sacramental Soil Moisture Accounting, Sac-SMA; \cite{anderson2005sacramento}) on CONUS-wide rainfall-runoff simulation tasks. Secondly, the spatial distribution of S4D-FT and LSTM performance across all study watersheds are illustrated to compare the model performance at different geographic locations. Lastly, a detailed investigation is conducted to analyze the possible factors driving regional variability in S4D-FT’s performance across CONUS. Based on our three-step evaluation, we conclude that the S4D-FT model demonstrates overall better performance and identify the conditions under which S4D-FT outperforms or underperforms the LSTM model.

\section{Datasets and Methodology}
\label{sec2}
In this study, we train and test a basic S4D, a variant S4D with frequency tunning (i.e., S4D-FT), and LSTM for rainfall-runoff simulations. The methodologies for S4D and S4D-FT are described in \textcolor{blue}{Section \ref{sec2.1}}. The LSTM follows the implementation of \cite{kratzert2019toward,kratzert2019towards} to match state-of-the-art benchmarks. The hyperparameters of S4D and S4D-FT are manually tunned through a trial-and-error process (details provided in \textcolor{blue}{Supplementary Tables \ref{tab:lstm_params} and \ref{tab:ssm_params}}).  

The training setup of the employed DL models aligns with existing LSTM and Transformer benchmarks ~\cite{frame2022deep,liu2024probing}. Specifically, the DL models are trained using long-term hydrometeorological time series and catchment attributes of 531 unimpaired watersheds across CONUS from the Catchment Attributes and MEteorology for Large-sample Studies (CAMELS; \citealp{addor2017camels,newman2015development}) as well as the corresponding streamflow measurements from the United States Geological Survey (USGS). In consistent with previous studies, we use 32 input variables (5 hydrometeorological variables from North American Land Data Assimilation System (NLDAS; \citealp{xia2012continental}) and 27 static catchment attributes, detailed in Supplementary Table \ref{tab:input_training_variables}) from 10/1/1999 to 9/30/2008 for training and 10/1/1989 to 9/30/1999 for testing. We train each DL model with different random seeds to form an eight-member ensemble for robustness. All DL models are trained with the Adam optimizer using a sequence-to-one approach and a 365-day look-back window. 

Statistical performance of the employed DL models is evaluated at each study watersheds using six evaluation metrics, including Pearson-r correlation, Nash-Sutcliffe Efficiency (NSE; \citealp{nash1970river}), Kling-Gupta Efficiency (KGE; \citealp{gupta2009decomposition}), percent bias in flow duration curve high-segment (top 2\%) volume (FHV; \citealp{yilmaz2008process}), percent bias in flow duration curve low-segment (lowest 30\%) volume (FLV; \citealp{yilmaz2008process}), and the overall percentage bias (PBias). The mathematical formulations of the employed evaluation statistics are provided in Supplementary Text \textcolor{blue}{S1} and Table \ref{tab:evaluation_metrics}. Notably, we also include statistical values of other popular DL models and a physical Sac-SMA model from \cite{frame2022deep} and \cite{liu2024probing} for a comprehensive comparison. 

Additionally, we analyze conditions under which S4D-FT outperforms or underperforms LSTM by linking model performance to specific statistical measures and hydrologic signatures. Details of the experiment design are provided in Section \ref{sec2.2}.

\subsection{State Space Models for rainfall-runoff simulations}
\label{sec2.1}
State Space Models (SSMs)~\citep{gu2021efficiently, gu2021combining} have recently gained increasing attention due to their strong performance in handling long-sequence data, and provide a promising alternative even to modern recurrent neural networks (RNNs)~\citep{ruschlong,erichson2021lipschitz,erichson2022gated}. SSMs’ architecture enables them to efficiently model complex sequential data while addressing some of the computational and stability challenges commonly faced by recurrent neural networks. This makes SSMs well-suited for hydrologic applications, such as rainfall-runoff modeling, where long-term dependencies and evolving temporal patterns are critical.

The foundation of SSMs is built upon continuous-time linear time-invariant (LTI) systems, which provide a structured approach for capturing relationships between inputs, outputs, and latent states over time. These relationships can be represented by the following equations:

\begin{equation}\label{eq.contLTI} 
\begin{aligned} \mathbf{x}'(t) &= \mathbf{A} \mathbf{x}(t) + \mathbf{B} \mathbf{u}(t), \ \mathbf{y}(t) &= \mathbf{C} \mathbf{x}(t) + \mathbf{D} \mathbf{u}(t), \end{aligned} 
\end{equation} 
where $\mathbf{u}(t) \in \C^{m}$, $\mathbf{y}(t) \in \C^{p}$, and $\mathbf{x}(t) \in \C^{n}$ are the inputs, outputs, and latent states, respectively. Here, the matrices $\mathbf{A} \in \C^{n \times n}$, $\mathbf{B} \in \C^{n \times m}$, $\mathbf{C} \in \C^{p \times n}$, and $\mathbf{D} \in \C^{m \times p}$ are the trainable parameters. 
Each matrix serves a distinct role in the model: $\mathbf{A}$ defines how the state evolves over time, $\mathbf{B}$ determines how inputs influence the state, $\mathbf{C}$ maps the state to the output, and $\mathbf{D}$ directly relates inputs to outputs. 
This continuous formulation is later discretized for practical implementation, using a trainable sampling interval $\Delta t > 0$, to yield the following discrete LTI system to process data sequentially over time:
\begin{equation}\label{eq.discreteLTI}
\begin{aligned}
\mathbf{x}_k &= \overline{\mathbf{A}} \mathbf{x}_{k-1} + \overline{\mathbf{B}} \mathbf{u}_k, \ \mathbf{y}_k &= \overline{\mathbf{C}} \mathbf{x}_k + \overline{\mathbf{D}} \mathbf{u}_k,
\end{aligned}
\end{equation}

Although LTI systems are linear, SSMs gain the ability to capture complex, nonlinear relationships by stacking multiple LTI systems and connecting them with nonlinear transformations, creating a deep model.  At first glance, the SSM structure may resemble an RNN, leading one to question its effectiveness. However, SSMs have several unique advantages that address key limitations of RNNs and LSTMs, particularly in handling long-range dependencies. In an RNN or LSTM, each time step relies on previous steps, often resulting in slow training and inference. SSMs, on the other hand, leverage the linear properties of the LTI system, which allows them to process sequences in parallel, either in the time domain~\citep{smith2023simplified} or frequency domain~\citep{yu2024robustifying,parnichkun2024state}. This parallelism not only speeds up training but also improves the model’s efficiency in handling long sequences. 

Additionally, SSMs avoid the well-known issues of exploding and vanishing gradients that plague RNNs. They achieve this by constraining the spectrum of the state matrix $\mathbf{A}$ to the left half of the complex plane~\citep{gu2021efficiently,gu2022parameterization}, thus stabilizing gradients, and by reparameterizing $\mathbf{A}$ and adopting small sampling intervals $\Delta t$ to prevent gradient decay~\citep{wang2023stablessm,yu2024there}.

In this study, we employ two specific variants of SSMs for rainfall-runoff simulations, termed Diagonal State Space Sequence (S4D) and Frequency Tuned S4D (S4D-FT). The S4D model simplifies the architecture by setting $\mathbf{A}$ to be diagonal and configuring the LTI system for single-input/single-output (SISO) operations, where $m = p = 1$. The S4D-FT is adopted from~\citet{yu2024tuning} which further rescales the imaginary part of $\mathbf{A}$ at initialization for the tuning of intrinsic bias that comes from the distribution of the eigenvalues of $\mathbf{A}$ in the Laplace domain. Though numerous other SSM variants exist~\citep{hasani2022liquid,smith2023simplified,agarwal2023spectral,yu2024there},~\citet{yu2024tuning} shows that the S4D models remain highly competitive when the frequency bias~\citep{basri,yu2022tuning} is tuned by multiplying $\text{Im}(\mathbf{A})$ by a hyperparameter $\alpha > 0$ at initialization.

\subsection{Attribution analyses of S4D-FT's relative performance over LSTM}
\label{sec2.2}
To understand in what situations S4D-FT outperforms or underperforms LSTM, we divided the study watersheds into two groups for further attribution analysis. Group 1 includes watersheds where S4D-FT consistently outperforms LSTM, with both positive NSE and KGE skill scores. Group 2 includes the remaining watersheds with negative NSE and/or KGE skill scores, suggesting that S4D-FT does not completely outperform LSTM.

The analysis includes two parts. First, we identify which statistical aspects of S4D-FT’s simulation drive NSE and KGE improvements or deteriorations. We compute correlations between NSE and KGE skill scores and improvements in additional evaluation metrics (FHV, Pearson-r, and PBias) across 531 study watersheds. To further validate our findings, we present simulated and observed hydrographs from the testing period for two representative watersheds from each group (i.e., one where S4D-FT performs well and one poorly). Formulations for computing FHV, Pearson-r, and PBias improvements are in \textcolor{blue}{Supplementary Table \ref{tab:evaluation_metrics}}.

Second, we investigate how watershed characteristics influence S4D-FT’s performance. We select eight hydrologic signatures as indicators of watershed streamflow characteristics, including mean daily discharge (q\_mean), 5\% flow quantile (low flow; q5), 95\% flow quantile (high flow; q95), frequency of high-flow days (high\_q\_freq), average duration of high-flow events (high\_q\_dur), frequency of low-flow days (low\_q\_freq), average duration of low-flow events (low\_q\_dur), and frequency of no-flow days (zero\_q\_freq). Technical descriptions of each hydrologic signature are in \textcolor{blue}{Supplementary Table \ref{tab:hydrologic_signatures}}. Further, we calculate percentage differences in these signatures between Groups 1 and 2 and assess their correlation with NSE and KGE skill scores. Formulations for percentage differences are in \textcolor{blue}{Supplementary Table \ref{tab:evaluation_metrics}}. 

\section{Results}
\label{sec3}
\subsection{Overall Performance of SSMs and Existing Benchmarks}
\label{sec3.1}
The statistical performance of S4D-FT, basic S4D, and other benchmarks (Sac-SMA, LSTM, MC-LSTM, Transformers, and Modified Transformers) for rainfall-runoff simulations across CONUS is summarized in Table \ref{tab:statistics}. Median values of evaluation metrics are presented for all study watersheds, with standard deviations across ensemble members shown in parentheses where available. The results for MC-LSTM, Transformers, Modified Transformers and Sac-SMA are directly adopted from previous studies. The results for S4D-FT, the basic S4D, and LSTM are produced by our own simulation experiments. Notably, our LSTM results align with benchmarks reported in the literature~\citep{frame2022deep, kratzert2019toward,kratzert2019towards,liu2024probing}.

According to Table \ref{tab:statistics}, S4D-FT demonstrates the best median NSE, KGE, Pearson-r, and FHV among all models. Moreover, S4D-FT shows the lowest standard deviation in NSE and Pearson-r, indicating greater consistency across different random seed initializations. In terms of FLV, the Transformers (including the basic Transformers and the Modified Transformers) outperform Sac-SMA, LSTM-type models (i.e., LSTM and MC-LSTM), and SSMs (i.e. S4D and S4D-FT). Regarding the overall bias (PBias), Sac-SMA still achieves the best performance, which is closest to zero, followed by LSTM-type models, and then SSMs. 

Comparing the basic S4D with S4D-FT, the basic S4D achieves only slightly better accuracy than the basic Transformers but still underperforms the LSTM and Modified Transformers. However, frequency tuning (i.e., S4D-FT) notably enhances S4D’s performance, improving all metrics except for FLV and establishing S4D-FT as the overall best-performing model among all existing benchmarks.

\begin{table}[h]
\centering
\caption{Statistical performance comparison of models using various metrics. Each metric value represents the median across 531 watersheds, with standard deviations calculated from different ensemble members shown in parentheses. Values without parentheses indicate that the standard deviation is unavailable. ``N/A'' denotes a metric not provided in previous studies. Metrics marked with ``\(\uparrow\)'' indicate that higher values are preferred, while ``\(\to 0\)'' means that values closer to zero are optimal. Best-performing metrics (medians closest to ideal) are highlighted in red.}

\label{tab:statistics}
\vspace{+0.1cm}
\resizebox{\textwidth}{!}{%
\begin{tabular}{>{\centering\arraybackslash}p{2.5cm}|cccccc}
\toprule
\textbf{Model} & \textbf{NSE}~(\(\uparrow\)) & \textbf{KGE}~(\(\uparrow\)) & \textbf{Pearson-r}~(\(\uparrow\)) & \textbf{FHV (\%)}~(\(\to 0\)) & \textbf{FLV (\%)}~(\(\to 0\)) & \textbf{PBias (\%)}~(\(\to 0\)) \\
\midrule
Sac-SMA & $0.65 \, (\pm 0.004)$ & $0.66 \, (\pm 0.006)$ & $0.82 \, (\pm 0.001)$ & $-21.36 \, (\pm 0.47)$ & $38.46 \, (\pm 2.31)$ & \textcolor{red}{\textbf{$2.53 \, (\pm 0.38)$}} \\
\midrule
LSTM & $0.72 \, (\pm 0.005)$ & $0.74 \, (\pm 0.007)$ & $0.86 \, (\pm 0.002)$ & $-17.51 \, (\pm 1.17)$ & $10.63 \, (\pm 6.18)$ & $5.42 \, (\pm 1.34)$ \\
\midrule
MC-LSTM\textsuperscript{1} & $0.72$ & $0.72$ & $0.86$ & $-18.72$ & $-30.84$ & $5.02$ \\
\midrule
Transformers\textsuperscript{2} & N/A & $0.71 \, (\pm 0.007)$ & N/A & $-26.66 \, (\pm 2.83)$ & $3.31 \, (\pm 2.34)$ & N/A \\
\midrule
Modified Transformers\textsuperscript{2} & N/A & $0.74 \, (\pm 0.007)$ & N/A & $-18.00 \, (\pm 2.94)$ & \textcolor{red}{\textbf{$2.28 \, (\pm 4.24)$}} & N/A \\
\midrule
S4D & $0.72 \, (\pm 0.004)$ & $0.72 \, (\pm 0.03)$ & $0.86 \, (\pm 0.001)$ & $-18.07 \, (\pm 4.13)$ & $16.18 \, (\pm 23.50)$ & $5.92 \, (\pm 8.83)$ \\
\midrule
S4D-FT & \textcolor{red}{\textbf{$0.74 \, (\pm 0.002)$}} & \textcolor{red}{\textbf{$0.75 \, (\pm 0.019)$}} & \textcolor{red}{\textbf{$0.87 \, (\pm 0.001)$}} & \textcolor{red}{\textbf{$-16.98 \, (\pm 2.26)$}} & \textbf{$20.17 \, (\pm 20.77)$} & \textbf{$5.87 \, (\pm 3.31)$} \\
\bottomrule
\end{tabular}%
}
\vspace{0.3cm}
\begin{minipage}{\textwidth}
\footnotesize
\textsuperscript{1} Results adopted from \cite{frame2022deep}.\\
\textsuperscript{2} Results adopted from \cite{liu2024probing}.
\end{minipage}
\end{table}

\subsection{Regional Performance Comparison of S4D-FT and LSTM}
\label{sec3.2}
\begin{figure}[!h]
\noindent\includegraphics[width=\textwidth]{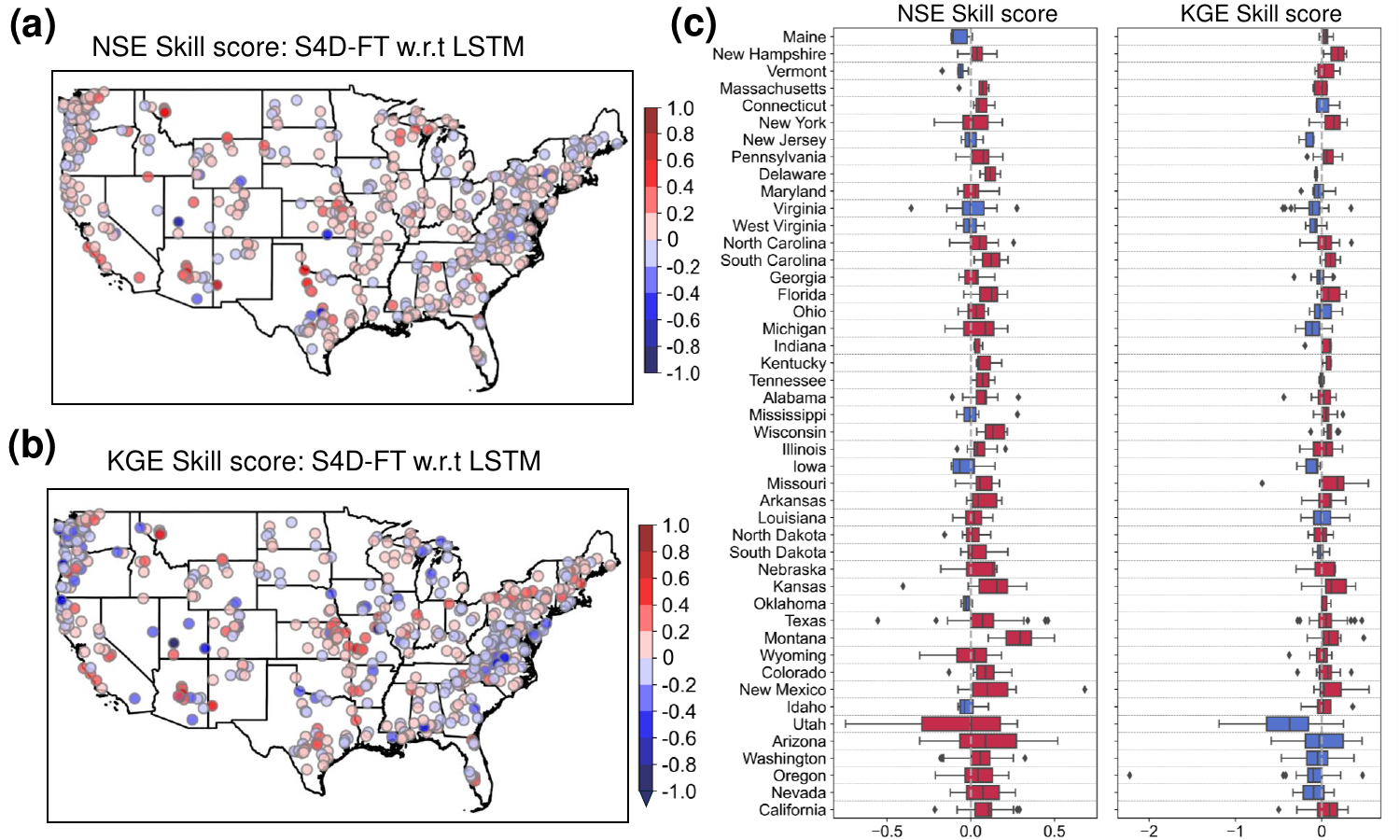}
\caption{Simulation performance of S4D-FT relative to the LSTM model across study watersheds in CONUS. Panel a: Spatial distribution of NSE skill scores, with red dots (positive NSE skill score) indicating S4D-FT outperformance and blue dots (negative NSE skill score) indicating LSTM outperformance. Panel b: Spatial distribution of KGE skill scores, following the same color scheme as Panel a. Panel c: Boxplots of NSE and KGE skill scores by state, ordered east to west, with red boxes for positive median scores and blue for negative medians. Rhode Island and Minnesota are excluded due to no study watersheds.}
\label{fig:composite_figure_1}
\end{figure}

Since LSTM is recognized as the leading model for CONUS-wide rainfall-runoff simulations, we focus on comparing S4D-FT and LSTM in regional performance. Figure \ref{fig:composite_figure_1}(a) and \ref{fig:composite_figure_1}(b) show NSE and KGE skill scores that demonstrate the relative improvement of S4D-FT over the baseline LSTM. Positive skill scores (red) indicate better performance by S4D-FT, while negative scores (blue) indicate LSTM outperforms S4D-FT. A skill score of 1 reflects theoretical best performance of S4D-FT (NSE or KGE = 1), and a skill score of 0 denotes equal performance between LSTM and S4D-FT. Darker colors represent greater differences.

According to Figure \ref{fig:composite_figure_1}(a) and \ref{fig:composite_figure_1}(b), S4D-FT presents better NSE and KGE prevalently across much of the CONUS, as shown by the widespread distribution of red dots. The better performance of S4D-FT is particularly pronounced in the Pacific Southwest and Mid-South (including Kansas, Missouri, Arkansas, and Texas). However, S4D-FT underperforms LSTM in several regions with different spatial patterns between NSE and KGE skill scores. Specifically, NSE skill scores show clusters of negative values (blue markers) along the East Coast (e.g., Maine and Virginia) and scattered negative values in parts of the Midwest. In contrast, KGE skill scores display more frequent negative values, particularly along the East Coast (e.g., Virginia), Great Lakes, Midwest (e.g., Utah), and Pacific Northwest (Washington and Oregon).

A more detailed breakdown of NSE and KGE skill scores by U.S. state is provided in Figure \ref{fig:composite_figure_1}(c), where red boxes represent states with a positive median skill score for NSE or KGE, and blue boxes indicate a negative median. According to Figure \ref{fig:composite_figure_1}(c), S4D-FT outperforms LSTM in most states, confirming the observations from 
 Figure \ref{fig:composite_figure_1}(a) and \ref{fig:composite_figure_1}(b). However, S4D-FT underperforms LSTM along the East Coast (specifically New Jersey, Virginia, and West Virginia) and in the Great Lakes region (specifically Minnesota and Iowa). For KGE skill scores alone, S4D-FT also shows weaker performance in the western U.S., particularly in Utah, Arizona, Washington, Oregon, and Nevada.

\subsection{Investigating factors behind regional variability in S4D-FT performance}
\label{sec3.3}
\begin{figure}[!h]
\noindent\includegraphics[width=\textwidth]{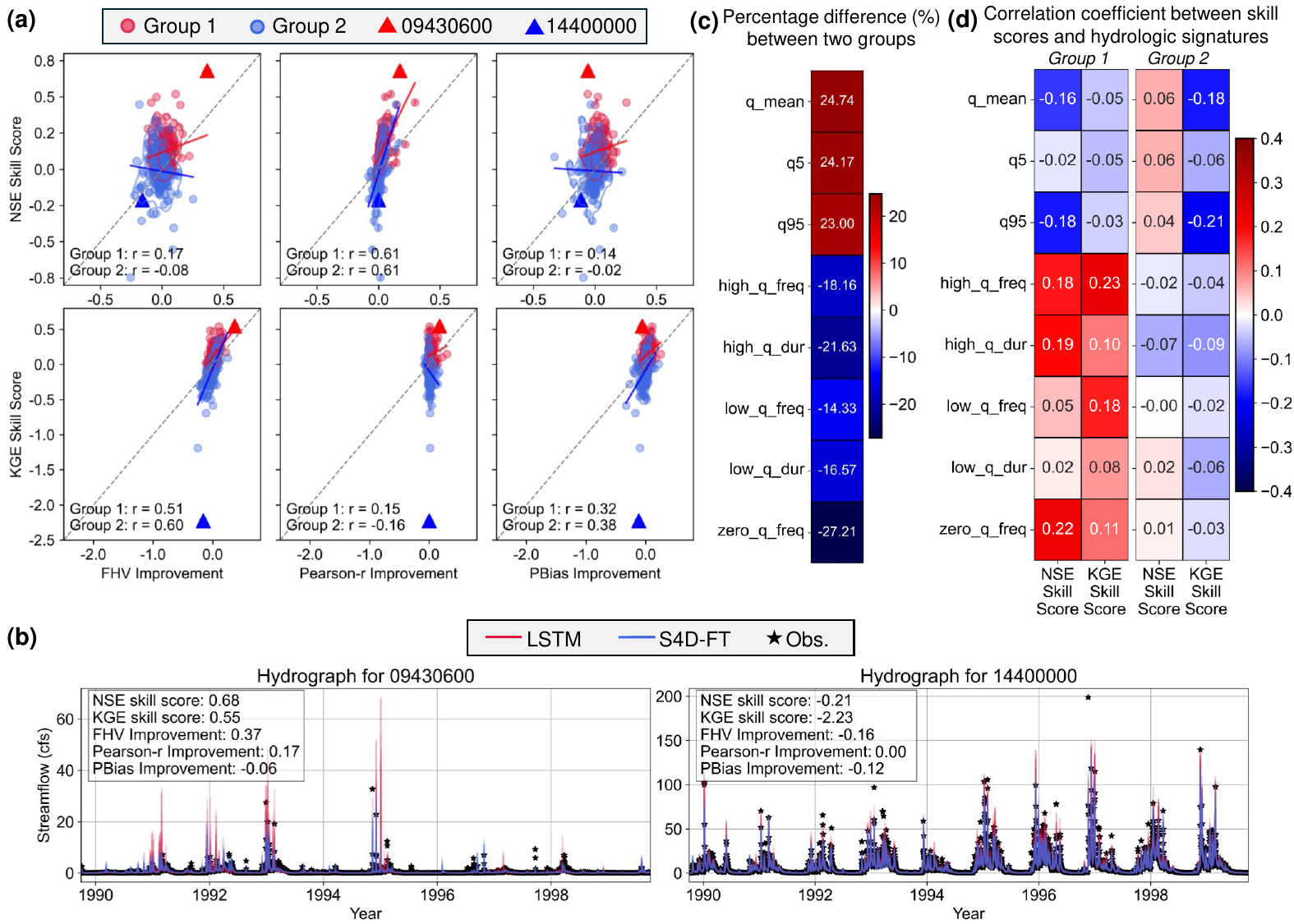}
\caption{Analysis of S4D-FT’s performance relative to LSTM considering multiple evaluation statistics and hydrological signatures. Panel a: Scatter plots overlaid with contour density plots of NSE and KGE skill scores against improvements in FHV, Pearson correlation, and PBias for Group 1 (red) and Group 2 (blue) watersheds. Solid red and blue lines represent regression lines. Correlation coefficients are displayed at the bottom left of each plot. The red and blue triangles highlight two specific example watersheds (i.e., 09430600 and 14400000) from Group 1 and Group 2, respectively. Panel b: Simulated 8-member ensemble hydrographs for LSTM (red) and S4D-FT (blue), along with observed streamflow (black stars), for the highlighted watersheds (i.e., 09430600 and 14400000). Panel c: heatmaps of percentage differences in the eight selected hydrologic signatures between Group 1 and Group 2 watersheds (only$> \pm 10\%$ differences shown) with values labeled in each cell. Panel d: heatmaps of correlation coefficients between the selected hydrologic signatures and NSE/KGE skill scores for Group 1 and Group 2 watersheds, with values labeled in each cell.}
\label{fig:composite_figure_2}
\end{figure}

To further investigate the varying performance of the S4D-FT over LSTM, we divide all study watersheds into two groups based on the relative performance of S4D-FT over LSTM (described in \textcolor{blue}{Section \ref{sec2.2}}). Figure \ref{fig:composite_figure_2}(a) presents scatter plots overlaid with contour density of S4D-FT’s NSE or KGE skill scores against its improvements or deterioration in FHV, Pearson-r, and PBias for the two groups of watersheds (Group 1 in red, Group 2 in blue). According to Figure \ref{fig:composite_figure_2}(a), NSE skill scores strongly correlate with Pearson-r improvements (r = 0.61) across both groups but have limited correlation with improvements in FHV and PBias, especially for Group 2 watersheds. In contrast, for KGE skill scores, both groups of watersheds exhibit the strongest positive correlation with FHV improvements (r = 0.51 and r = 0.60 for Groups 1 and 2, respectively), followed by PBias improvements (r = 0.32 and r = 0.38 for Groups 1 and 2, respectively), but weak correlations with Pearson-r improvements.

Figure \ref{fig:composite_figure_2}(b) displays the simulated and observed hydrographs for two representative watersheds: one with the highest KGE skill score (USGS station 09430600, red triangle in Figure \ref{fig:composite_figure_2}(a)) and one with the lowest KGE skill score (USGS station 14400000, blue triangle in Figure \ref{fig:composite_figure_2}(a)). According to Figure \ref{fig:composite_figure_2}(b), at watershed 09430600, S4D-FT alleviates the LSTM’s significant overestimation and improves FHV (by 0.37) and Pearson-r (by 0.17). However, S4D-FT does not present better PBias. Conversely, at watershed 14400000, both LSTM and S4D-FT underestimate high flows, with S4D-FT showing more severe underestimation, leading to decreased FHV (by -0.16) and PBias (by -0.12). Pearson-r at watershed 14400000 remains unchanged (improvement by 0), which aligns with the weak relationship observed in Figure \ref{fig:composite_figure_2}(a) between Pearson-r improvement and KGE skill scores. 

In addition, we examine the relationship between NSE or KGE skill scores and watershed characteristics. Figure \ref{fig:composite_figure_2}(c) presents a heatmap of percentage differences in eight hydrologic signatures between Groups 1 and 2 watersheds. Red cells indicate higher hydrologic signature values in Group 2 watersheds, and blue cells indicate the opposite. Darker colors represent greater differences. According to Figure \ref{fig:composite_figure_2}(c), Group 2 watersheds tend to have higher streamflow volumes (q\_mean, q5, q95), as indicated by positive percentage differences. In contrast, Group 2 watersheds exhibit lower frequency and shorter duration of both high and low flow events (high\_q\_freq, high\_q\_dur, low\_q\_freq, low\_q\_dur, and zero\_q\_freq), as shown by negative percentage differences.

Lastly, Figure \ref{fig:composite_figure_2}(d) provides the correlation coefficients between NSE or KGE skill scores and eight hydrologic signatures for both Group 1 and Group 2 watersheds. Negative correlations are shown in blue, positive in red, with darker colors indicating stronger correlations (positive or negative). According to Figure \ref{fig:composite_figure_2}(d), for Group 1 watersheds, higher q\_mean and q95 are associated with worse NSE skill scores, as demonstrated by the negative correlations. However, S4D-FT tends to have better NSE skill scores in watersheds with more frequent and prolonged high-flow and no-flow events, as suggested by positive correlations with high\_q\_freq, high\_q\_dur, and zero\_q\_freq. In terms of the KGE in Group 1 watersheds, S4D-FT performs better in watersheds with more frequent high and low flows, while no significant relationships exist with the remaining hydrologic signatures. In contrast, for Group 2 watersheds, S4D-FT’s NSE skill scores show no significant correlations with any hydrologic signatures. In terms of KGE, relatively stronger negative correlations with q\_mean and q95 suggest watersheds with higher streamflow volumes may favor LSTM over S4D-FT for rainfall-runoff simulations.

\section{Discussion}
\label{sec4}
The growing use of DL models in hydrology highlights the need for a standardized evaluation framework to better understand their strengths and limitations. Current studies often differ in training and testing dataset, study periods, geospatial regions, and/or the number of watersheds analyzed, making direct comparisons difficult and potentially misleading. Therefore, we advocate adopting a standardized simulation setup (hereafter referred to as “the standard setup”) following \cite{frame2023strictly} and \cite{liu2024probing}, which uses 531 CAMELS watersheds with NLDAS forcing, training from 10/1/1999–9/30/2008, and testing from 10/1/1989–9/30/1999. 

Under the standard setup, the decades-old LSTM remains the top performer and prevailing benchmark in rainfall-runoff simulations, outperforming even newer architectures like Transformers. This has led to speculation that LSTM may have already reached, or is approaching, the predictive limit for rainfall-runoff simulations \citep{liu2024probing}. In an effort to push the boundaries of simulation accuracy, we carry out this first-of-its-kind research in introducing the latest SSMs, specifically the S4D-FT, to the hydrology community, exploring its capability in advancing rainfall-runoff simulations, comparing its performance to the existing prevailing DL benchmark, and seeking to establish a new benchmark for this task.

Our results show a favorable performance of S4D-FT as compared to the LSTM with the overall median NSE value increased from 0.72 to 0.74, and the median KGE value increased from 0.74 to 0.75 over a total of 531 watersheds across CONUS. Moreover, S4D-FT outperforms LSTM consistently across different regions in CONUS, with exceptions in a few areas such as the eastern U.S. (e.g., West Virginia and Virginia), the Great Lakes (e.g., Minnesota and Iowa), and Pacific Northwest considering both NSE and KGE (Figure \ref{fig:composite_figure_1}). 

Interestingly, S4D-FT shows greater improvements in NSE than KGE across more regions (Figure \ref{fig:composite_figure_1}). Given that both NSE and KGE are composite metrics, such discrepancy leads to two additional scientific questions: (1) Why does S4D-FT underperform on KGE more frequently than on NSE? (2) And does this suggest specific strengths or limitations of S4D-FT in rainfall-runoff simulation compared to LSTM? Our results in Figure \ref{fig:composite_figure_2}(a) and \ref{fig:composite_figure_2}(b) reveal that S4D-FT’s KGE performance is strongly correlated with high-flow regime bias (FHV) but shows only a minimal correlation with temporal correspondence (Pearson-r correlation). In contrast, NSE is less affected by high-flow regime bias and is primarily driven by temporal consistency with observations. Although S4D-FT outperforms LSTM in both FHV and Pearson-r in terms of the overall statistics (Table \ref{tab:statistics}), our further analysis indicates that S4D-FT results in a higher proportion of watersheds with improved Pearson-r compared to FHV (75\% and 50\% of the study watersheds, respectively according to \textcolor{blue}{Supplementary Table \ref{tab:performance_ratios}}). As a result, the FHV-sensitive nature of KGE leads to fewer improvements compared to the correlation-sensitive NSE. Given such information, we suspect that S4D-FT’s primary strength over LSTM lies in its capability of capturing temporal correspondence. However, S4D-FT may offer limited improvement in simulating high-flow regimes compared to LSTM.

Further analysis of the relationship between streamflow characteristics and S4D-FT’s NSE or KGE skill scores (Figure \ref{fig:composite_figure_2}(c) and \ref{fig:composite_figure_2}(d)) show that S4D-FT outperforms LSTM in watersheds satisfying the following conditions at the same time: more frequent and prolonged high-flow, low-flow, and zero-flow events, as well as smaller flow magnitudes. These conditions could be seen in watersheds with snowmelt-driven streamflow regimes, such as those in the Rocky Mountains and Sierra Nevada  \citep{addor2017camels,brunner2020future,pham2021evaluation,yang2023classification}, where streamflow follows a seasonal pattern with low flows during snow accumulation and high flows during snowmelt. Similar conditions could also be found in watersheds with intermittent streamflow regimes characterized by high evapotranspiration and short precipitation events (e.g., thunderstorms or fronts). These watersheds are located primarily in the Great Plains (from North Dakota to Texas, as well as Missouri and Arkansas) \citep{addor2017camels,brunner2020future,yang2023classification}. Notably, the above-mentioned regions align closely with the study locations where S4D-FT outperforms LSTM (Figure \ref{fig:composite_figure_1}).

Conversely, we further identify that there is a subset of watersheds where S4D-FT underperforms LSTM in KGE, notably clustered along the East Coast and West Coast (Figure \ref{fig:composite_figure_1}). Such an underperformance is primarily linked to large daily mean streamflow and high-flow magnitudes (Figure \ref{fig:composite_figure_2}(d), Group 2 heatmap), which reinforces our earlier suspect of S4D-FT’s limited capability in simulating high-flow volumes (i.e., relatively less improved FHV). Furthermore, the geographic distribution of negative KGE skill scores aligns with the locations of watersheds characterized by large flow volumes, particularly in the Mid-Atlantic along the East Coast and the Pacific Northwest (Figure \ref{fig:composite_figure_1}). These areas are characterized by pluvio-nival streamflow regimes (i.e., a combination of rainfall and snowmelt) that feature infrequent but intense high-flow events  \citep{addor2017camels,yang2023classification}.

While the improved accuracy of S4D-FT represents advancements in applying DL models to hydrologic simulations, its black-box nature remains a limitation. Our current explanation of S4D-FT's success, though plausible, remains superficial, as it does not fully uncover the specific processes that S4D-FT excels at or struggles with in rainfall-runoff simulation. We believe this challenge is not unique to S4D-FT; rather, it is a recurring question for all data-driven DL models in hydrology: \textit{What specific hydrologic processes are these models effectively simulating, and where do they fall short? }

This challenge arises in part from the differing priorities between the computer science (CS) and hydrology communities. While the CS community prioritizes achieving higher predictive accuracy, the hydrology community focuses equally, if not more, on model interpretability. For hydrologists, a model’s value is not only determined by its statistical performance but also by its ability to provide insight into the mechanisms driving certain hydrologic phenomena. Although it is reported that the examination of physical-meaningful hidden states of DL could also provide the underneath physical insights \citep{lees2021hydrological}, we note such an approach may not be applied universally as the identification of the physical-meaningful hidden node(s) is not guaranteed.

Alternatively, we believe further advancing physics-aware DL models is a promising direction. By integrating physical principles into data-driven model structures, physics-aware DL models could strike a balance between purely physics-based and purely data-driven approaches to achieve both interpretability and high simulation accuracy. A noteworthy example is the Mass Conserving (MC) LSTM \citep{hoedt2021mc}, which imposes a strict mass conservation constraint on the standard LSTM structure. It is reported that MC-LSTM achieves significantly higher accuracy than physically-based models while providing greater interpretability than conventional LSTM \citep{frame2022deep}. Such an explicit link between MC-LSTM’s model states and real-world hydrologic components enables the analysis of MC-LSTM’s latent representations, thereby offering interpretation from the perspective of fundamental water balance principles.

Despite theoretical advantages and promising results reported thus far, we acknowledge that physics-aware DL may come at the cost of some predictive accuracy \cite{frame2022deep}). Nevertheless, we argue that the trade-off between accuracy and interpretability shall be considered worthwhile if it enables hydrologists to trace model predictions back to specific physical processes, deepening our understanding of hydrologic systems and providing insights beyond pure statistical analysis. Looking ahead, we envision DL models being tailored to the needs of the hydrology community to achieve a better balance between performance and interpretability following significant progress made thus far \citep{feng2022differentiable,hou2024physics,ji2024groundwater,shen2023differentiable,tsai2021calibration}.

\section{Conclusions}
\label{sec5}
This study proposes adopting a first-of-kind S4D-FT model for rainfall-runoff simulations of a total of 531 watersheds in CONUS. Through a comprehensive evaluation of statistical metrics and spatial performance, we demonstrate that S4D-FT outperforms the current leading model, i.e., the decades-old LSTM model, in large-scale rainfall-runoff simulations.

Our analysis highlights that S4D-FT excels in watersheds where high- and low-flow events are frequent and prolonged, but with smaller high- and low-flow magnitudes among all study watersheds. These watersheds are often associated with snowmelt-driven regimes, such as those in the Rocky Mountains, and intermittent flow-dominated regions, like parts of the Great Plains. However, S4D-FT tends to be less effective in regions with high daily mean and peak streamflow values, such as the pluvio-nival watersheds in the Mid-Atlantic and Pacific Northwest. The limited performance of S4D-FT might be associated with less accurate simulations of high-flow regimes.

To conclude, our findings show that S4D-FT transcends the predictive limits of LSTM in rainfall-runoff simulations over a large number of study cases, demonstrating that S4D-FT sets a new community-based standard for CONUS-wide rainfall-runoff simulation tasks using DL models. By introducing S4D-FT to the hydrologic community and benchmarking its capabilities in rainfall-runoff simulations, we advocate for the broader use of S4D-FT in hydrologic applications as a more advanced alternative to the conventional LSTM model.  

Future research should explore the integration of generative models for probabilistic time series forecasting~\cite{zhou2023deep,naimangenerative,lim2024elucidating} in hydrologic applications. These models can provide uncertainty estimates, which is critical for decision-making in water resource management and flood risk assessment. By capturing the inherent variability and uncertainty in rainfall-runoff processes, probabilistic approaches can provide more robust predictions, particularly in extreme flow scenarios. Combining the predictive power of SSMs like S4D-FT with generative frameworks could pave the way for next-generation hydrologic forecasting tools, offering both enhanced accuracy and actionable uncertainty estimates.

\section*{Acknowledgment}
The University of Oklahoma (OU) team acknowledges support of the National Science Foundation (NSF) CAREER Award (No. 2236926). The OU team would also like to thank the Department of Defense, Army Corps of Engineers (DOD-COR) Engineering With Nature (EWN) Program (Award No. W912HZ-21-2-0038), the U.S. Bureau of Reclamation (USBR) Project No. R24AC00032, National Oceanic and Atmospheric Administration (NOAA) 's Climate Program Office, CVP and MAPP programs (Award Number: NA23OAR4310459), and the NSF Grant No. OIA-1946093 and its subaward No. EPSCoR-2020-3. AY would like to thank the SciAI Center, funded by the Office of Naval Research under Grant Number N00014-23-1-2729. NBE would like to acknowledge partial support from the U.S. Department of Energy, Office of Science, Office of Advanced Scientific Computing Research, SciDAC program, under Contract Number DE-AC02-05CH11231; and NSF under Grant No. 2319621.

\section*{Open Research}
The training and testing data (hydrometeorological variables, static catchment attributes, and USGS gauge streamflow) is from the Catchment Attributes and Meteorology for Large-sample Studies (CAMELS) dataset, available at \url{https://gdex.ucar.edu/dataset/camels.html}. The Sacramento Soil Moisture Accounting (Sac-SMA) with SNOW-17 simulations are adopted from \url{https://www.hydroshare.org/resource/d750278db868447dbd252a8c5431affd/}. The Python codes to reproduce the results of this paper is available at \url{https://github.com/WESTENR-OU/S4D_rainfall_runoff_simulations}. 

\bibliography{references}
\bibliographystyle{iclr2025_conference}

\clearpage
\appendix
\section*{Appendix}
\section{Text S1: Evaluation strategy}
The performance of Sac-SMA, LSTM, and S4D-FT are evaluated from two perspectives. Firstly, the overall statistical performance of three models (along with other DL benchmarks including MC-LSTM, Transformers, and basic S4D) is comprehensively evaluated using six statistical metrics. These metrics include the Pearson-r correlation, Nash-Sutcliffe Efficiency (NSE; \citealp{nash1970river}), Kling-Gupta Efficiency (KGE; \citealp{gupta2009decomposition}), percent bias in flow duration curve high-segment (top 2\%) volume (FHV; \citealp{yilmaz2008process}), percent bias in flow duration curve low-segment (lowest 30\%) volume (FLV; \citealp{yilmaz2008process}), as well as the overall percentage bias (PBias). The detailed formulation of each metric is provided in \textcolor{blue}{Supplementary Table \ref{tab:evaluation_metrics}}. 

Secondly, we present a spatial distribution of S4D-FT’s performance compared to LSTM for a clearer and more direct comparison for watersheds at different geospatial locations. We employ NSE and KGE skill scores to illustrate the relative accuracy between S4D-FT and LSTM. Further details on the computation of NSE and KGE skill scores can also be found in \textcolor{blue}{Supplementary Table \ref{tab:evaluation_metrics}}. We do not conduct spatial performance analysis for the remaining DL models or the physically-based Sac-SMA, as these models have been shown to perform less effectively than the LSTM \citep{frame2022deep,kratzert2019toward,kratzert2019towards,liu2024probing}.

\newpage
\section{Input variables for SSMs and LSTM training}
\renewcommand{\thetable}{S1}
\begin{table}[!ht]
\centering
\caption{Hydrometeorological variables and static catchment attributes for DL training. Variable descriptions are adopted from \cite{kratzert2019toward}.}
\resizebox{0.87\textwidth}{!}{%
\begin{tabular}{p{0.14\textwidth}p{0.035\textwidth}p{0.2\textwidth}p{0.475\textwidth}}
\hline
\textbf{Category} & \textbf{No.} & \textbf{Variable} & \textbf{Description} \\ \hline
\multirow{5}{=}{\centering \parbox{0.14\textwidth}{Hydro-\\meteorological\\Variables}}
    & 1  & Maximum air temp  & 2 m daily maximum air temperature (\(^\circ\text{C}\)) \\
    & 2  & Minimum air temp  & 2 m daily minimum air temperature (\(^\circ\text{C}\)) \\
    & 3  & Precipitation     & Average daily precipitation (mm/day) \\
    & 4  & Radiation         & Surface-incident solar radiation (W/m\(^2\)) \\
    & 5  & Vapor pressure    & Near-surface daily average vapor pressure (Pa) \\ \hline
\multirow{27}{=}{\centering Static Catchment Attributes} 
    & 6  & Precipitation mean      & Mean daily precipitation \\
    & 7  & PET mean                & Mean daily potential evapotranspiration \\
    & 8  & Aridity index           & Ratio of mean PET to mean precipitation \\
    & 9  & Precip seasonality      & Annual precipitation and temperature represented as sine waves. Positive (negative) values indicate precipitation peaks during summer (winter). Values near 0 suggest uniform precipitation throughout the year. \\
    & 10 & Snow fraction           & Fraction of precipitation on days with temperature \(< 0^\circ\text{C}\) \\
    & 11 & High precipitation frequency & Frequency of days with precipitation \( \geq 5 \times \) mean daily precipitation \\
    & 12 & High precip duration    & Average duration of high precipitation events (number of consecutive days with \( \geq 5 \times \) mean daily precipitation) \\
    & 13 & Low precip frequency    & Frequency of dry days (\( < 1 \) mm/day) \\
    & 14 & Low precip duration     & Average duration of dry periods (consecutive days with precipitation \( < 1 \) mm/day) \\
    & 15 & Elevation               & Mean elevation of the catchment \\
    & 16 & Slope                   & Mean slope of the catchment \\
    & 17 & Area                    & Catchment area \\
    & 18 & Forest fraction         & Fraction of catchment covered by forest \\
    & 19 & LAI max                 & Maximum monthly mean of leaf area index \\
    & 20 & LAI difference          & Difference between the maximum and minimum mean leaf area index \\
    & 21 & GVF max                 & Maximum monthly mean of green vegetation fraction \\
    & 22 & GVF difference          & Difference between the maximum and minimum monthly mean green vegetation fraction \\
    & 23 & Soil depth (Pelletier)  & Depth to bedrock (maximum 50 m) \\
    & 24 & Soil depth (STATSGO)    & Soil depth (maximum 1.5 m) \\
    & 25 & Soil porosity           & Volumetric soil porosity \\
    & 26 & Soil conductivity       & Saturated hydraulic conductivity \\
    & 27 & Max water content       & Maximum soil water content \\
    & 28 & Sand fraction           & Fraction of sand in the soil \\
    & 29 & Silt fraction           & Fraction of silt in the soil \\
    & 30 & Clay fraction           & Fraction of clay in the soil \\
    & 31 & Carbonate rocks fraction & Fraction of catchment area classified as carbonate sedimentary rocks \\
    & 32 & Geological permeability & Surface permeability (log\(_{10}\)) \\ \hline
\end{tabular}%
}
\label{tab:input_training_variables}
\end{table}

\newpage
\section{Model Structure and Training Hyperparameters}
\renewcommand{\thetable}{S2}
\begin{table}[ht]
\centering
\caption{LSTM training hyperparameters}
\begin{tabular}{|c|c|}
\hline
\textbf{Parameter}           & \textbf{Value} \\ \hline
hidden\_size                 & 256            \\ \hline
batch\_size                  & 256            \\ \hline
initial\_forget\_gate\_bias  & 3              \\ \hline
learning\_rate               & 1e-3 for 1-10 epochs, 5e-4 for 11-25 epochs, and 1e-4 for 26-30 epochs          \\ \hline
dropout                      & 4e-1            \\ \hline
epochs                       & 30             \\ \hline
\end{tabular}
\label{tab:lstm_params}
\end{table}

\renewcommand{\thetable}{S3}
\begin{table}[ht]
\centering
\caption{SSMs (basic S4D and S4D-FT) training hyperparameters}
\begin{tabular}{|c|c|}
\hline
\textbf{Parameter}           & \textbf{Value} \\ \hline
d\_model                     & 128            \\ \hline
d\_state                     & 128            \\ \hline
n\_layer                     & 6              \\ \hline
cfi (only for S4D-FT)                          & 10             \\ \hline
cfr (only for S4D-FT)                    & 10             \\ \hline
ssm\_dropout                 & 1.2e-1           \\ \hline
epochs                       & 50             \\ \hline
epochs\_scheduler            & 50             \\ \hline
batch\_size                  & 128            \\ \hline
lr                           & 4e-4         \\ \hline
lr\_min                      & 4e-5        \\ \hline
lr\_dt                       & 1e-3          \\ \hline
min\_dt                      & 1e-2           \\ \hline
max\_dt                      & 1e-1            \\ \hline
weight\_decay                & 3e-2          \\ \hline
wd                           & 2e-2           \\ \hline
\end{tabular}
\label{tab:ssm_params}
\end{table}

\newpage
\section{Evaluation Metrics}
\renewcommand{\thetable}{S4}
\begin{table}[!ht]
\centering
\caption{Evaluation metrics and their formulations, ranges, and optimal values.}
\renewcommand{\arraystretch}{1.7} 
\resizebox{\textwidth}{!}{
\begin{tabular}{|>{\normalsize\raggedright}m{4.6cm}|>{\large\centering\arraybackslash}m{10.5cm}|>{\normalsize\centering\arraybackslash}m{1.6cm}|>{\normalsize\centering\arraybackslash\normalsize}m{1.7cm}|}
\hline
\textbf{Metric} & \textbf{Formulation} & \textbf{Range} & \textbf{Optimum} \\ \hline

\textbf{Pearson-r} & 
$\frac{\sum_{i=1}^n (Q_{\text{sim},i}-\overline{Q}_{\text{sim}})(Q_{\text{obs},i}-\overline{Q}_{\text{obs}})}{\sqrt{\sum_{i=1}^n (Q_{\text{sim},i}-\overline{Q}_{\text{sim}})^2 \sum_{i=1}^n (Q_{\text{obs},i}-\overline{Q}_{\text{obs}})^2}}$ 
& $(-\infty, 1]$ & $1$ \\ \hline

\textbf{Nash-Sutcliffe Efficiency (NSE)} & 
$1 - \frac{\sum_{i=1}^n (Q_{\text{obs},i}-Q_{\text{sim},i})^2}{\sum_{i=1}^n (Q_{\text{obs},i}-\overline{Q}_{\text{obs}})^2}$ 
& $(-\infty, 1]$ & $1$ \\ \hline

\textbf{Kling-Gupta Efficiency (KGE)} & 
$1 - \sqrt{(CC-1)^2 + \left(\frac{\overline{Q}_{\text{sim}}}{\overline{Q}_{\text{obs}}} - 1\right)^2 + \left(\frac{\sigma_{\text{sim}}}{\sigma_{\text{obs}}} - 1\right)^2}$ 
& $(-\infty, 1]$ & $1$ \\ \hline

\textbf{Percent bias in flow duration curve high-segment volume (FHV)} & 
$\frac{\sum_{h=1}^H (Q_{\text{sim},h}-Q_{\text{obs},h})}{\sum_{h=1}^H Q_{\text{obs},h}} \times 100$ 
& $(-\infty, \infty)$ & $0$ \\ \hline

\textbf{Percent bias in flow duration curve low-segment volume (FLV)} & 
$\frac{-\sum_{l=1}^L \left[\log(Q_{\text{sim},l})-\log(Q_{\text{sim},L})\right] \, \left(-\sum_{l=1}^L \left[\log(Q_{\text{obs},l})-\log(Q_{\text{obs},L})\right]\right)}{\sum_{l=1}^L \left[\log(Q_{\text{obs},l})-\log(Q_{\text{obs},L})\right]} \times 100$ 
& $(-\infty, \infty)$ & $0$ \\ \hline

\textbf{Percentage Bias (PBias)} & 
$\frac{\sum_{i=1}^n Q_{\text{sim},i} - \sum_{i=1}^n Q_{\text{obs},i}}{\sum_{i=1}^n Q_{\text{obs},i}} \times 100$ 
& $(-\infty, \infty)$ & $0$ \\ \hline

\textbf{NSE skill score (model relative to reference)} & 
$\frac{\text{NSE}_{\text{model}} - \text{NSE}_{\text{ref}}}{1 - \text{NSE}_{\text{ref}}}$
& $(-\infty, 1]$ & $1$ \\ \hline

\textbf{KGE skill score (model relative to reference)} & 
$\frac{\text{KGE}_{\text{model}} - \text{KGE}_{\text{ref}}}{1 - \text{KGE}_{\text{ref}}}$
& $(-\infty, 1]$ & $1$ \\ \hline

\textbf{FHV improvement (model relative to reference)} & 
$\frac{\left| FHV_{\text{ref}} - 1 \right| - \left| FHV_{\text{model}} - 1 \right|}{100}$ 
& $(-\infty, \infty)$ & $\infty$ \\ \hline

\textbf{Pearson-r improvement (model relative to reference)} & 
$\left| Pearsonr_{\text{ref}} - 1 \right| - \left| Pearsonr_{\text{model}} - 1 \right|$ 
& $(-\infty, \infty)$ & $\infty$ \\ \hline

\textbf{PBias improvement (model relative to reference)} & 
$\frac{\left| PBias_{\text{ref}} \right| - \left| PBias_{\text{model}} \right|}{100}$ 
& $(-\infty, \infty)$ & $\infty$ \\ \hline

\textbf{Percentage difference in hydrologic signature $h$ between watersheds Group 2 and Group 1} & 
$\frac{\overline{h_{\text{Group 2}}} - \overline{h_{\text{Group 1}}}}{\overline{h_{\text{Group 1}}}} \times 100$ 
& $(-\infty, \infty)$ & $0$ \\ \hline

\end{tabular}
}
\label{tab:evaluation_metrics}
\end{table}

\vspace{0.25cm} 
\textbf{Notations:} \\
$n$: Length of the testing sample time series. \\
$H$: Number of flow indices that fall within the exceedance probability of 0.02 (i.e., top 2\% of flow volumes). \\
$L$: Number of flow indices that fall within the 30\% low-flow segment (i.e., 0.7-1.0 exceedance probability range). \\

\newpage
\section{Description of hydrologic signatures}
\renewcommand{\thetable}{S5}
\begin{table}[!h]
\centering
\caption{Description of employed hydrologic signatures. This table is partially adopted from \cite{addor2017camels}.}
\begin{tabular}{|>{\centering\arraybackslash}m{3cm}|>{\centering\arraybackslash}m{9cm}|>{\centering\arraybackslash}m{2cm}|}
\hline
\textbf{Hydrologic Signature} & \textbf{Description} & \textbf{Unit} \\ \hline
q\_mean & Mean daily discharge & mm/day \\ \hline
q5 & 5\% flow quantile (low flow) & mm/day \\ \hline
q95 & 95\% flow quantile (high flow) & mm/day \\ \hline
high\_q\_freq & Frequency of high-flow days (\textgreater 9 times the median daily flow) & days/year \\ \hline
high\_q\_dur & Average duration of high-flow events (number of consecutive days \textgreater 9 times the median daily flow) & days \\ \hline
low\_q\_freq & Frequency of low-flow days (\textless 0.2 times the mean daily flow) & days/year \\ \hline
low\_q\_dur & Average duration of low-flow events (number of consecutive days \textless 0.2 times the mean daily flow) & days \\ \hline
zero\_q\_freq & Frequency of days with Q = 0 & \% \\ \hline
\end{tabular}
\label{tab:hydrologic_signatures}
\end{table}

\newpage
\section{Supplementary Results}
\renewcommand{\thetable}{S6}
\begin{table}[h!]
\centering
\caption{Percentage of watersheds with improved (first row) and decreased (second row) performance of S4D-FT across various metrics, including NSE, KGE, Pearson-r, FHV, FLV, and PBias.}
\begin{tabular}{|l|c|c|c|c|c|c|}
\hline
\textbf{Ratio (\%)} & \textbf{NSE} & \textbf{KGE} & \textbf{Pearson-r} & \textbf{FHV} & \textbf{FLV} & \textbf{PBias} \\ \hline
Improved & 68.7 & 54.4 & 75.0 & 49.5 & 51.4 & 49.7 \\ \hline
Decreased & 31.3 & 45.6 & 25.0 & 50.5 & 48.6 & 50.3 \\ \hline
\end{tabular}
\label{tab:performance_ratios}
\end{table}

\end{document}